\documentclass[11pt,a4paper]{article}
\usepackage[hyperref]{emnlp2018}
\usepackage{times}
\usepackage{latexsym}
\usepackage{url}
\usepackage{textcomp}
\usepackage{bbm}
\usepackage{amsmath}
\usepackage{booktabs}
\usepackage{tabularx}
\usepackage{graphicx}
\usepackage{dialogue}
\usepackage{mathtools}
\usepackage{hyperref}

\usepackage{multirow}
\usepackage{mdframed}
\usepackage{tcolorbox}

\usepackage{xcolor,pifont}

\newcommand\cmark {\textcolor{green}{\ding{52}}}
\newcommand\xmark {\textcolor{red}{\ding{55}}}
\mdfdefinestyle{dialogue}{
    backgroundcolor=yellow!20,
    innermargin=5pt
}

\usepackage{amssymb}
\usepackage{soul}
\makeatletter
\def\blfootnote{\xdef\@thefnmark{}\@footnotetext}
\makeatother

\aclfinalcopy % Uncomment this line for the final submission

\newcommand{\ours}[0]{\daffy}%\textsc{quac}}
\newcommand{\ourstext}[0]{QuAC}

\newcommand{\daffy}[0]{\includegraphics[width=.04\textwidth]{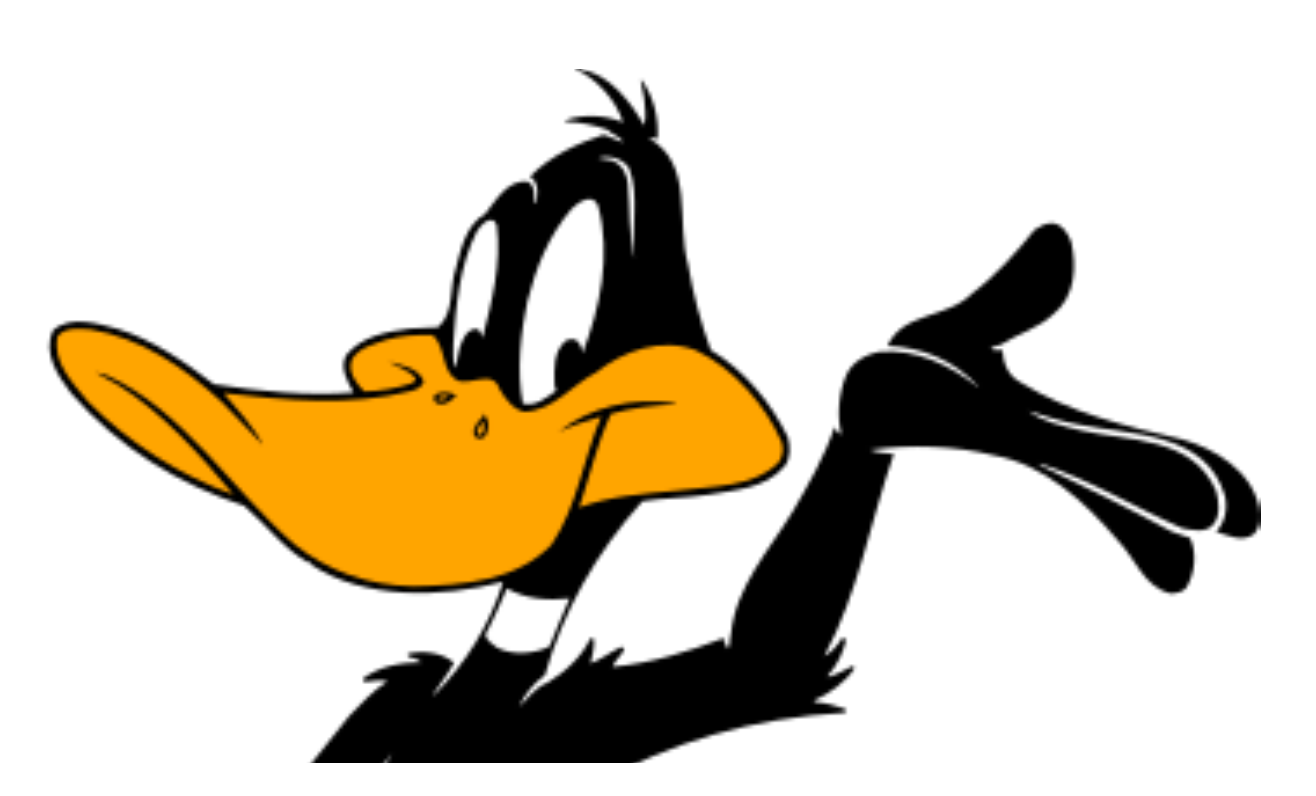}}

\newcolumntype{L}[1]{>{\raggedright\let\newline\\\arraybackslash\hspace{0pt}}m{#1}}
\newcolumntype{C}[1]{>{\centering\let\newline\\\arraybackslash\hspace{0pt}}m{#1}}
\newcolumntype{R}[1]{>{\raggedleft\let\newline\\\arraybackslash\hspace{0pt}}m{#1}}

\title{ \daffy~QuAC  : Question Answering in Context}

\author{
\textbf{
Eunsol Choi$^{\bigstar\heartsuit}$ \hspace{5mm} He He$^{\bigstar\diamondsuit}$ \hspace{5mm} Mohit Iyyer$^{\bigstar\clubsuit\dagger}$ \hspace{6mm} Mark Yatskar$^{\bigstar\dagger}$} \\
\textbf{
Wen-tau Yih$^{\dagger}$ \hspace{4mm} Yejin Choi$^{\heartsuit\dagger}$ \hspace{4mm} Percy Liang$^{\diamondsuit}$ \hspace{4mm} Luke Zettlemoyer$^{\heartsuit}$} \vspace{2mm} \\ 
Allen Institute for Artificial Intelligence$^{\dagger}$ \hspace{4mm} University of Washington$^{\heartsuit}$\\ Stanford University$^{\diamondsuit}$ \hspace{5mm} UMass Amherst$^{\clubsuit}$ 
\vspace{2mm} \\
\texttt{\{eunsol,yejin,lsz\}@cs.washington.edu} \\ \texttt{\{hehe,pliang\}@cs.stanford.edu}\\ \texttt{\{mohiti,marky,scottyih\}@allenai.org}
}

\date{}

\begin{document}
\maketitle
\begin{abstract}

We present \ourstext, a dataset for \textbf{Qu}estion \textbf{A}nswering in \textbf{C}ontext that contains 14K information-seeking QA dialogs (100K questions in total). The dialogs involve two crowd workers: (1) a student who poses a sequence of freeform questions to learn as much as possible about a hidden Wikipedia text, and (2) a teacher who answers the questions by providing short excerpts from the text. 
\ourstext\ introduces challenges not found in existing machine comprehension datasets: its questions are often more open-ended, unanswerable, or only meaningful within the dialog context, as we show in a detailed qualitative evaluation.
We also report results for a number of reference models, including a recently state-of-the-art reading comprehension architecture extended to model dialog context. 
Our best model underperforms humans by 20 F1, suggesting that there is significant room for future work on this data.
%We also introduce a strong neural baseline that considers both the reference text and the dialog context but still underperforms humans by 23.8 F1, suggesting that there is significant room for future work on this data.
%that \ours\ is a useful dataset for learning more effective information-seeking dialog agents.
Dataset, baseline, and leaderboard available at \url{http://quac.ai}.
\end{abstract}
\begin{table*}[t]
 \small
 \centering
% \begin{tabular}{|p{2 cm}|p{1 cm}|p{1 cm}|p{1 cm}|p{1 cm}|p{1 cm}|} \\ \hline
% \begin{tabular}{|c|c|p{1.8cm}|c|p{2.2 cm}|p{2cm}|}
\begin{tabular}
{c|c|c|c|c|c|c}
%{|c|c|p{2cm}|c|p{2.3cm}|p{2cm}|p{2 cm}|}
\toprule
\multirow{2}{*}{Dataset}& \multirow{2}{*}{\parbox{1cm}{\centering Multi turn}} &\multirow{2}{*}{\parbox{1cm}{\centering Text-based}} & \multirow{2}{*}{\parbox{1.2cm}{\centering Dialog Acts}} &
\multirow{2}{*}{\parbox{1.5 cm}{\centering Simple Evaluation}}  & \multirow{2}{*}{\parbox{1.7cm}{\centering Unanswerable Questions}}  & \multirow{2}{*}{\parbox{2cm}{\centering Asker Can't See Evidence}} \\
&&&&&\\\midrule
\textbf{\daffy~QuAC} & \cmark&\cmark &   \cmark & \cmark & \cmark & \cmark  \\%Web and Wikipedia\\
\midrule
CoQA~\cite{Reddy18} &\cmark &\cmark & \xmark & \cmark & \cmark & \xmark \ \\
CSQA~\cite{saha2018complex} &\cmark &\xmark & \xmark & \xmark & \cmark & \xmark \ \\
CQA~\cite{talmor2018web} &  \cmark&\cmark & \xmark & \cmark & \xmark & \cmark \\
SQA~\cite{IyyerSQA2017}&\cmark& \xmark & \xmark & \cmark & \xmark & \xmark \\ 
NarrativeQA~\cite{Kocisk2017TheNR} & \xmark& \cmark & \xmark & \xmark & \xmark & \cmark \\
TriviaQA~\cite{JoshiTriviaQA2017}& \xmark & \cmark & \xmark & \cmark & \xmark & \cmark\\
SQuAD 2.0~\cite{Rajpurkar2018KnowWY} & \xmark& \cmark &  \xmark &  \cmark & \cmark & \xmark\\%Wikipedia\\
MS Marco~\cite{Nguyen2016MSMA} & \xmark &\cmark & \xmark & \xmark & \cmark & \cmark\\
NewsQA~\cite{trischler2016newsqa} & \xmark&\cmark & \xmark &  \cmark & \cmark & \cmark \\
\bottomrule 
\end{tabular}
\caption{Comparison of the QUAC dataset to other question answering datasets.} 
\vspace{-10pt}
 \label{tab:dataset_comparison}
\end{table*}

\begin{figure}[t]
%\vspace{-50pt}
%\footnotesize
% \begin{center}
%\colorbox{Goldenrod}
%\begin{mdframed}[style=dialogue]
\footnotesize

\begin{tcolorbox}[boxsep=0pt,
                  left=5pt,
                  right=0pt,
                  top=2pt,
                  colback = yellow!5
                  ]%%

    % \hspace{1.25in}
    % \includegraphics[width=.08\textwidth]{figures/daffy_duck.jpg} \\
    \hspace{15pt}{\textbf{Section}:
    \includegraphics[width=.08\textwidth]{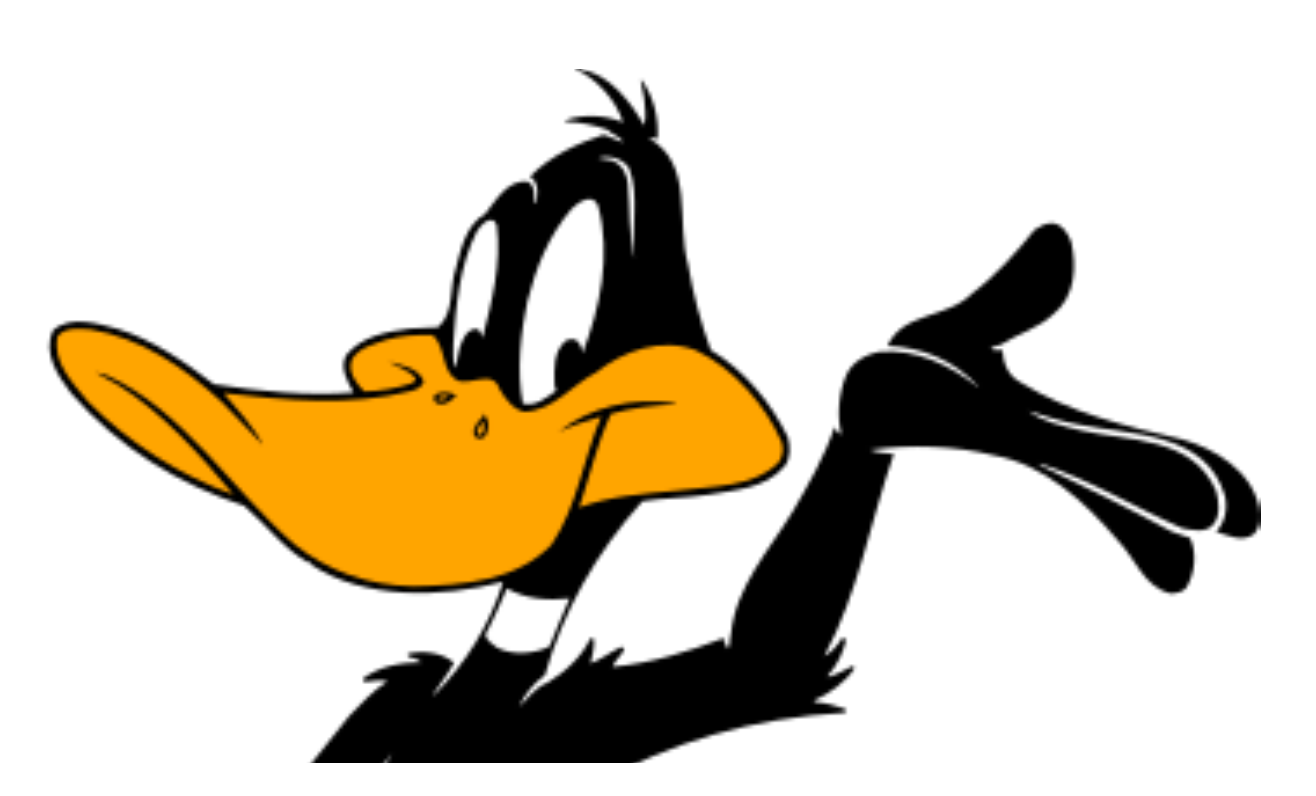}\href{https://en.wikipedia.org/wiki/Daffy_Duck\#Origin_and_history}{ Daffy Duck, Origin \& History}}\\

\begin{dialogue}

% \end{center}
\vspace{-5mm}

\speak{Student}{\bf What is the origin of Daffy Duck?}
\speak{Teacher}\colorbox{CarnationPink!25}{$\hookrightarrow$} %Daffy
{first appeared in Porky's Duck Hunt}%, released on April 17, 1937.
%Follow up.

 \speak{Student}{\bf What was he like in that episode?}
\speak{Teacher} \colorbox{CarnationPink!25}{$\hookrightarrow$}
{~assertive, unrestrained, combative}

\speak{Student}{\bf Was he the star?}
\speak{Teacher}\colorbox{CarnationPink!25}{$\bar{\hookrightarrow}$} \colorbox{red!25}{No,} {barely more than an unnamed bit player in this short}

%Follow up? If you want

\speak{Student}{\bf Who was the star?}
\speak{Teacher}\colorbox{CarnationPink!25}{ $\not\hookrightarrow$} \colorbox{blue!15}{No answer}

%Do not follow up

\speak{Student}{\bf Did he change a lot from that first episode in future episodes?}
\speak{Teacher}\colorbox{CarnationPink!25}{ $\hookrightarrow$} \colorbox{red!25}{Yes,}{the only aspects of the character that have remained consistent (...) %through the years
are his voice characterization by Mel Blanc}%; and his black feathers with a white neck ring.
%Follow up

\speak{Student}{\bf How has he changed?}
\speak{Teacher}\colorbox{CarnationPink!25}{ $\hookrightarrow$}
{Daffy was less anthropomorphic} %(...) % and resembles a ``normal'' black duck.
%Follow up, multiple answers

\speak{Student}{\bf In what other ways did he change?}
\speak{Teacher}\colorbox{CarnationPink!25}{ $\hookrightarrow$}
Daffy's slobbery, exaggerated lisp (...) is barely noticeable in the early cartoons.

\speak{Student}{\bf Why did they add the lisp? }
\speak{Teacher}\colorbox{CarnationPink!25}{ $\hookrightarrow$}
One often-repeated ``official'' story is that it was modeled after producer Leon Schlesinger's tendency to lisp.

%Follow up, multiple answers

\speak{Student}{\bf Is there an ``unofficial'' story? }
\speak{Teacher}\colorbox{CarnationPink!25}{ $\hookrightarrow$}
\colorbox{red!25}{Yes,} Mel Blanc (...) contradicts that conventional belief

$\ldots$

%Follow up

%\speak{Student}{\bf What's his position?}
%\speak{Teacher}\colorbox{CarnationPink!25}{ $\not\hookrightarrow$} 
%such an extended mandible would hinder his speech, particularly on words containing an \emph{s} sound.
%Do not follow up

%\vspace{-10pt}
\end{dialogue}
\end{tcolorbox}

%\end{mdframed}

\vspace{-10pt}
\caption{An example dialog about a Wikipedia section. The student, who does not see the section text, asks questions. 
The teacher provides a response in the form of a text span (or \colorbox{blue!15}{No answer}), optionally yes or no (\colorbox{red!25}{{\footnotesize Yes}} / \colorbox{red!25}{{\footnotesize No}}), and encouragement about continuing a line of questioning (should, \colorbox{CarnationPink!25}{$\hookrightarrow$}, could \colorbox{CarnationPink!25}{$ \bar\hookrightarrow$}, or should not \colorbox{CarnationPink!25}{$ \not\hookrightarrow$} ask a follow-up question).}

%. Formally, the teacher provides (1) whether the student should \colorbox{CarnationPink!25}{$\hookrightarrow$}, could \colorbox{CarnationPink!25}{$ \bar\hookrightarrow$}, or should not \colorbox{CarnationPink!25}{$ \not\hookrightarrow$} ask a follow-up; (2) affirmation (\colorbox{red!25}{{\footnotesize Yes}} / \colorbox{red!25}{{\footnotesize No}}), and, when appropriate, (3) \colorbox{blue!15}{No answer}.
%When the student's question is answerable from the section, the teacher provides spans of text. Several questions only make sense within the context of the dialog history.
%}
\label{fig:exdial}
\end{figure}

\section{Introduction}
\label{sec:introduction}

In information-seeking dialog, students repeatedly ask teachers questions to learn about a topic of interest~\cite{stede2004information}. Modeling such conversations is challenging, as the questions can be highly context-dependent, elliptical, and even unanswerable. To enable learning from rich information-seeking dialog, we present \ourstext~(henceforth \ours),
%\footnote{We refer to \ourstext~as ~\ours~and read it as \ourstext.}
a large-scale dataset for \textbf{Qu}estion \textbf{A}nswering in \textbf{C}ontext that contains 14K crowdsourced QA dialogs (100K total QA pairs).\footnote{We use ``dialog'' to refer to a sequence of QA pairs.}\blfootnote{$\bigstar$ \textnormal{Authors contributed equally.}}
%Most current datasets for reading comprehension (RC), which contain questions written about a given document, assume that the question writer has a high degree of familiarity with the contents of that document. In SQuAD~\cite{rajpurkar2016squad}, for example, the question writer has read the entire document beforehand and already knows the answer to their question. But what if they know nothing at all about the document or its contents? Users in this scenario are more likely to ask a series of questions rather than just a single one. The question writer (or \emph{student}) in Figure~\ref{fig:exdial} begins by asking for general information about the topic. As the interaction progresses, they latch onto details that interest them (e.g., Daffy Duck's lisp) and ask follow-up questions to learn more specific information.

% Figure~\ref{fig:exdial} shows an example \ours\ dialog. The interaction is student driven and centered around a short evidence text (a section from Daffy Duck's Wikipedia page), which only the teacher can access. Given just a section heading, ``Origin \& History'', the student aims to learn as much as possible about it by asking questions. The teacher answers these questions with spans from the evidence text and additional dialog acts. This formulation is closely related to recent work on machine comprehension of text (e.g., SQuAD~\cite{rajpurkar2016squad}) where known-answerable questions are posed in isolation. However, \ours\ adds the challenges of modeling open-ended, context-dependent, and unanswerable questions.

Figure~\ref{fig:exdial} shows an example \ours\ dialog.
% on a short evidence text. 
% (a section from Daffy Duck's Wikipedia page). 
% The teacher has access to the entire section, while students are given only the section heading (``Origin \& History''). Student leads the dialog to learn as much as possible about the section. % with the feedbacks from teacher. 
The interaction is student driven and centered around a short evidence text (a section from Daffy Duck's Wikipedia page), which only the teacher can access. Given just the section's heading, ``Origin \& History'', the student aims to learn as much as possible about its contents by asking questions. 
The teacher answers these questions with spans from the evidence text, as in existing reading comprehension tasks~\cite{rajpurkar2016squad}. Additionally, the teacher uses dialog acts to provide the student with feedback (e.g., ``ask a follow up question"), which makes the dialogs more productive. % This formulation is closely related to recent work on machine comprehension of text (e.g., SQuAD~\cite{rajpurkar2016squad}) where known-answerable questions are posed in isolation. 

%The types of questions asked by students engaged in these information-seeking interactions (or \emph{QA dialogs}) differ substantially from those asked by domain experts. In Figure~\ref{fig:exdial}, we observe highly contextual questions (e.g., ``what's his position?''), one of which is not even answerable from the document. To explore this complex, multi-turn RC setting, we crowdsource a large-scale dataset for \emph{question answering in context} (\ours), which contains roughly 14K QA dialogs (100K questions in total).

We collect the dataset in an interactive setting where two crowd workers play the roles of teacher and student. 
To encourage natural and diverse questions, we do not follow previous dialog-style QA datasets that semi-automatically generate questions~\cite{talmor2018web,saha2018complex}.
Furthermore, unlike QA datasets such as SQuAD and CoQA~\cite{Reddy18}, students in \ours\ do not know the answers to their questions prior to asking them, which lessens the role of string matching and simple paraphrasing in answering their questions. 
This property makes \ours\ similar to datasets that contain real user queries on search engines~\cite{Nguyen2016MSMA}. 
% they wish. 
% \pl{important: stress that questions where someone actually doesn't know the answer rather than knowing and quizzing the answerer;
% this is way more realistic as a dataset - more like user queries (MS MARCO) that you'd find on a search engine}
% \pl{another thing: maybe worth mentioning interesting aspects of the data collection process, like having two conversations going on at once}
% To make it easier for students to generate questions, we select the evidence text from a popular Wikipedia article about a person (or anthropomorphic fictional character). 
% Also, our pay scheme incentivizes workers to produce long dialogs covering most important content in the evidence. This interactive setting enables our dataset to contain open-ended, context-dependent, non-factoid, and unanswerable questions.

\ours\ contains many challenging phenomena unique to dialog, such as coreference to previous questions and answers and open-ended questions that must be answered without repeating previous information (Section~\ref{sec:analysis}). % and open-ended questions with multiple equally valid answers. 
%We find that student questions linearly track the evidence document, even though they do not have access to the text.
Additionally, despite lacking access to the section text, we find that students start dialogs by asking questions about the beginning of the section before progressing to asking questions about the end.
These observations imply that models built for \ours\ must incorporate the dialog context to achieve good performance. 
% As previous QA pairs give strong clue for the answer for next question (XX\% of answer comes from neighboring sentences of the prior answer), modeling dialogs will be crucial for this dataset. 
%Furthermore, XX of questions contain 
%We discover patterns in the way students navigate their way through the evidence document: the number of turns already completed in the dialog influences the location of the current question's answer within the evidence. These observations imply that models built for \ours\ must incorporate the dialog context to achieve good performance. 

We present a strong neural baseline~\cite{clark2017simple} that considers both dialog context and section text. While this model achieves within 6 F1 of human performance on SQuAD, it performs 20 F1 points below the human upper bound on \ours, indicating room for future improvement.

\section{Dataset collection}
\label{sec:dataset}
%This section describes how we use Mechanical Turk to collect \ours.

This section describes our data collection process, which involves facilitating QA dialogs between crowd workers.
Table~\ref{tab:dataset_comparison} shows \ours~shares many of the same positive characteristics of existing QA datasets while expanding upon the dialog aspect.
%: \ours\ is one of the first datasets of dialog-style QA on text.

%In contrast to existing multi-turn QA datasets, such as ATIS~\cite{dahl1994expanding} and SQA~\cite{IyyerSQA2017}, we are able create a large scale dataset with information-seeking behavior.
%, which motivates us to collect our own dataset. 
%This section describes how we address these short comings through the creation of \ours.

\begin{table}[t]
\vspace{5pt}
\small
    \centering
    \begin{tabular}{l|r|r|r|r}
    \toprule
    & Train & Dev. & Test & Overall \\ \midrule
    questions & 83,568 & 7,354&7,353 & 98,407\\
    dialogs & 11,567 &  1,000 & 1,002 & 13,594 \\%13,594\\
    unique sections & 6,843&1,000 & 1,002 & 8,854\\ %5,960 & 749 & 748 \\%8,854\\

  %  \# of unique answers & & & \\
    %\# of unanswerable questions & 14,459&1,487 &1,475\\
    %\# of yes/no questions & 22,030&1,622 & 1,720 \\
    %16,950\\
   % \# of articles &5,401 & 913 & 892 \\%3,611\\
    \midrule
     tokens / section & 396.8 & 440.0 & 445.8 & 401.0 \\
     tokens / question & 6.5 & 6.5 & 6.5  & 6.5  \\
     tokens / answer & 15.1 & 12.3 & 12.3 & 14.6 \\ % & & \\
     questions / dialog & 7.2 & 7.4& 7.3 & 7.2 \\
     \hspace{5pt}\% yes/no  & 26.4& 22.1& 23.4 & 25.8\\
     \hspace{5pt}\% unanswerable  & 20.2& 20.2& 20.1 &20.2\\
    \bottomrule
    \end{tabular}
    \caption{Statistics summarizing the \ours\ dataset. }
    \label{tab:stats}
\end{table}

\begin{figure*}[ht]
    \centering
    \vspace{-5pt}
    \includegraphics[width=\textwidth]{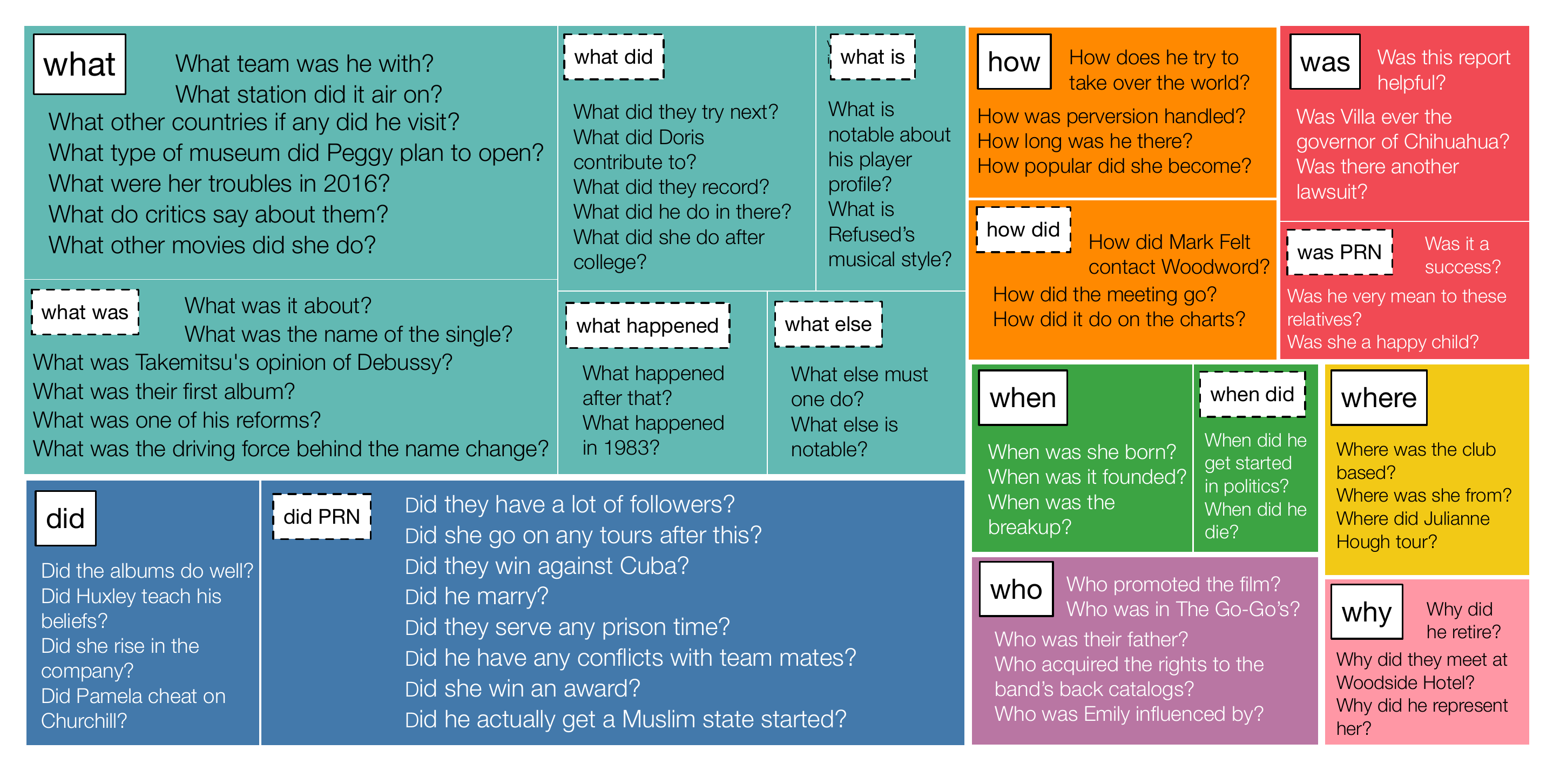}
    \vspace{-20pt}
    \caption{A treemap visualization of the eight most frequent ``Wh'' words in \ours, where box area is proportional to number of occurrences. Compared to other machine comprehension datasets, we observe increased contextuality and open-endedness, as well as a variety of both general and specific questions.}
    \label{fig:questions}
    \vspace{-5pt}
\end{figure*}

\subsection{Interactive Task}
Our task pairs up two workers, a \emph{teacher} and a \emph{student}, who discuss a section $s$ (e.g., \href{https://en.wikipedia.org/wiki/Daffy_Duck#Origin_and_history}{``Origin \& History''} in the example from Figure~\ref{fig:exdial}) from a Wikipedia article about an entity $e$ (Daffy Duck). 
The student is permitted to see only the section's title $t$ and the first paragraph of the main article $b$, while the teacher is additionally provided with full access to the section text. 
% \lzcomment{add explicit references to example in Fig 1 for this running example?}
% \pl{same idea: would be nice to see an example of $e$, $t$, $b$}

The task begins with the student formulating a free-text question $q$ from the limited information they have been given. 
The teacher is not allowed to answer with free text; instead, they must select a contiguous span of text defined by indices $(i,j)$ into the section text $s$.\footnote{We set the maximum answer length to 30 tokens to prevent teachers from revealing the full article all at once.}
While this decision limits the expressivity of answers, it makes evaluation simpler and more reliable; as such, it has been adopted in other reading comprehension datasets such as SQuAD, TriviaQA~\cite{JoshiTriviaQA2017}, and NewsQA~\cite{trischler2016newsqa}. 

To facilitate more natural interactions, teachers must also provide the student with a list of dialog acts $v$ that indicates the presence of any of $n$ discrete statements. 
% \pl{find the formality a bit pedantic}
We include three types of dialog acts: (1) continuation (\texttt{follow up}, \texttt{maybe follow up}, or \texttt{don't follow up}), (2) affirmation (\texttt{yes}, \texttt{no}, or \texttt{neither}) and (3) answerability (\texttt{answerable} or \texttt{no answer}). The continuation act is crucial for workers to have productive dialogs, as it allows teachers to guide the student's questioning towards aspects of the article that are especially important or interesting.
% \footnote{TODO : other marked bits that we ignore}    
% \pl{it's not totally clear at this point what each of these dialog acts means...
% that it's a request for the student to follow?}
Altogether, a teacher's complete answer to a question $q$ includes a pair of indices and dialog indicators, $a = (i, j, v)$.
If a question is marked \texttt{no answer}, the indices are ignored.

%and provide  or mark the question as unanswerable.
%Formally, 

%\footnote{We set the maximum answer span length to 30 tokens to prevent teachers from revealing the full article to the student at once.} While this decision limits the expressivity of the teacher's answers, it makes evaluation much simpler and more reliable; for this reason, other RC datasets such as SQuAD, TriviaQA~\cite{JoshiTriviaQA2017}, and NewsQA~\cite{trischler2016newsqa} also contain answer spans. 

%To reclaim some of the expressivity of free text, we allow the teacher to provide other bits of information in addition to the answer span: (1) whether the question is a yes/no question (and ``yes'' or ``no'' if it is), (2) whether the selected span directly answers the question or indirectly implies the answer, (3) whether the question has multiple valid answer spans within the section or just a single one, and (4) whether the student should ask a follow-up question or pursue a new line of questioning. 

After receiving an answer from the teacher, the student asks another question. At every turn, the student has more information about the topic than they %\pl{should I just give in to the fact that 'they' is the new gender neutral singular third-person pronoun?} 
did previously, which encourages them to ask follow-up questions about what they have just learned. 
The dialog continues until (1) twelve questions are answered, (2) one of the partners decides to end the interaction, or (3) more than two unanswerable questions were asked. %Each completed dialog consists of a sequence of question answer pairs $c =[ (q,a)_0 ... (q,a)_L]$.

%for a sequence of at most $L$ \pl{what's $L$? don't need since you give the stopping condition below} question-answer pairs to form a complete dialog, $c =[ (q,a)_0 ... (q,a)_L]$.
%In practice, we stop dialogs 

\subsection{Collection Details}
We used Amazon Mechanical Turk for collection, restricting the task to workers in English-speaking countries and with more than 1000 HITs with at least a 95\% acceptance rate.
We paid workers per the number of completed turns in the dialog, which encourages workers to have long dialogs with their partners, and discarded dialogs with less than three QA pairs.\footnote{On average, we paid \$0.33 per question, increasing pay per question as dialogs got longer to encourage completion.} 
To ensure quality, we created a qualification task and allowed workers to report their partner for various problems. More details on data collection can be found in our datasheet.\footnote{~\url{http://quac.ai/datasheet.pdf}}%\footnote{We also periodically disqualified workers who consistently participated in short dialogs.} 

\paragraph{Article selection}
Our early pilot studies showed that articles about people generally require less background knowledge to write good questions than other categories.
To find articles about people with varied backgrounds, we retrieved articles from a list of category keywords (culture, animal, people associated with event, geography, health, celebrity) using a web interface provided by the Wikimedia foundation.\footnote{https://petscan.wmflabs.org/} We pruned by popularity by selecting articles with at least 100 incoming links, and we additionally removed non-person entities using YAGO~\cite{Suchanek2007YagoAC}. After article selection, we filtered sections from these articles based on the number of paragraphs, number of tokens, and average words per sentence.
\footnote{These filtering steps bias our data towards entertainers; see datasheet for details.}
% : 36\% of the articles are about musicians, 15\% athletes, 13\% movie or television personalities, and the remaining 36\% an assortment of historical, scientific and political figures.}

\paragraph{Dataset validation}
To create our evaluation sets, we collected four additional annotations per question. 
Workers were presented with questions from a previously collected dialog and asked to provide answer spans.\footnote{After submitting an answer, they were shown the original teacher's answer so that they could understand the context of the subsequent questions.}
Acquiring many annotations is important since many questions in \ours\ have multiple valid answers. 

\paragraph{Train / Dev / Test Differences}
Table~\ref{tab:stats} shows small differences between training, development and testing splits. Sections in the training set are shorter than those in the evaluation folds because we permit multiple dialogs about the same section only in training; since workers preferred reading shorter sections, these were more likely to result in multiple dialogs.
Variations in answer span length arise from two sources: (1) having multiple annotations in the validation task and (2) differing incentives between the  data collection and validation procedures.\footnote{Validation workers did not have to maintain the dialog and so did not include as much information in the response.}
An analysis measuring the effect of these variations shows that they result in little difference in evaluation.\footnote{More specifically, we analyze whether references from the initial data collection significantly differ from references collected during validation. We observe a difference of less than 1 F1 when using the original answer as system output versus using validation answers.}

%\eunsol{todo: explain slight difference b/w validation and train}

%from the search results. 

%culture	animal	event	geography	health	tv	sports	celebrity
% min_intro_words = 100
% good_sec_word_len = 250
% max_sec_word_len = 550
% min_sent_len = 10
% max_sent_len = 50
% min_para_len = 2
% max_para_len = 5

\begin{figure}[t]
\small
% \begin{center}
%\colorbox{Goldenrod}
%\begin{mdframed}[style=dialogue]

\begin{tcolorbox}[boxsep=0pt,
                  left=5pt,
                  right=0pt,
                  top=2pt,
                  colback = yellow!5
                  ]%%

    % \hspace{1.25in}
    % \includegraphics[width=.08\textwidth]{figures/daffy_duck.jpg} \\
    \hspace{15pt}{\textbf{Section}: %{\includegraphics[width=.08\textwidth]{figures/pdf/daffyhand.pdf} 
    \href{https://en.wikipedia.org/wiki/Augusto_Pinochet#Intellectual_life_and_academic_work}{Augusto Pinochet : Intellectual life...}}\\

\begin{dialogue}

% \end{center}
\vspace{-5mm}
\small

\speak{Student}{\bf Was he known for being intelligent?}
\speak{Teacher}\colorbox{CarnationPink!25}{$\hookrightarrow$} %Daffy
\colorbox{red!25}{No,}{ Pinochet was publicly known as a man with a lack of culture.}%, released on April 17, 1937.
%Follow up.

 \speak{Student}{\bf why did people feel that way?}
\speak{Teacher} \colorbox{CarnationPink!25}{$\hookrightarrow$}{reinforced by the fact that he also portrayed himself as a common man}% with simple ideas.}

\speak{Student}{\bf did he have any hobbies?}
\speak{Teacher}\colorbox{CarnationPink!25}{${\hookrightarrow}$} \colorbox{red!25}{Yes,} {Before wresting power from Allende, Pinochet had written two books.}

%Follow up? If you want

\speak{Student}{\bf what is the name of a book written by him?}
\speak{Teacher}\colorbox{CarnationPink!25}{$\hookrightarrow$} {Geopolitica (1968) and Campana de Tarapaca (1972). }

%Do not follow up

\speak{Student}{\bf what were the books about?}
\speak{Teacher}\colorbox{CarnationPink!25}{ $\hookrightarrow$} {Chile's military literature.}%; and his black feathers with a white neck ring.
%Follow up

\speak{Student}{\bf was there anything noteworthy regarding his books?}
\speak{Teacher}\colorbox{CarnationPink!25}{ $\hookrightarrow$}
\colorbox{red!25}{Yes,}{In Geopolitica Pinochet plagiarized (...) Gregorio Rodriguez Tascon} %(...) % and resembles a ``normal'' black duck.
%Follow up, multiple answers

\speak{Student}{\bf did he deny those allegations?}
\speak{Teacher}\colorbox{CarnationPink!25}{ $\not\hookrightarrow$}\colorbox{blue!15}{No answer}

%Follow up.

\speak{Student}{\bf what did he plagiarize in Geopolitica? }
\speak{Teacher}\colorbox{CarnationPink!25}{ $\bar{\hookrightarrow}$}
In Geopolitica Pinochet plagiarized (...) paragraphs from a 1949 presentation
%
%his mentor general 
%Gregorio Rodriguez Tascon %(...) %by using paragraphs from a 1949 conference presentation.
$\ldots$

%Follow up, multiple answers

% \speak{Student}{\bf what academic work did he complete? }
% \speak{Teacher}\colorbox{CarnationPink!25}{ $\bar{\hookrightarrow}$}
% In contrast to the two latter Pinochet was not an outstanding student

% %Follow up

% \speak{Student}{\bf what kind of work did he do?}
% \speak{Teacher}\colorbox{CarnationPink!25}{ $\bar{\hookrightarrow}$} 
% his persistence and interest in geopolitics made Rodriguez assume the role as his academic mentor.

% \speak{Student}{\bf did Pinochet have any jobs?}
% \speak{Teacher}\colorbox{CarnationPink!25}{ $\bar{\hookrightarrow}$} 
% \colorbox{red!25}{Yes,} Rodriguez granted Pinochet a slot as assistant lecturer (...)%in geopolitics and in geography.
% %Do not follow up

%\vspace{-10pt}
\end{dialogue}
\end{tcolorbox}

\caption{An example successful dialog from \ours. Questions build on each other and interesting aspects (e.g., plagiarism) are explored as they are discovered.} 
\label{fig:exdial2}
\end{figure}

\section{Dataset Analysis}
\label{sec:analysis}

\ours\ differs from other reading comprehension datasets due to our dialog-style collection process and the information asymmetry between teacher and student. 
In the following sections, we provide a qualitative analysis of the dataset in \ours\ that highlights challenging question types as well as the impact of the dialog context.

\paragraph{Question and answer types}
Table~\ref{tab:stats} shows dataset summary statistics.
%\ours\ contains a broader array of questions than existing reading comprehension datasets. 
\ours~ has long answers of 15 tokens on average compared to 3 for SQuAD, which is unsurprising as most SQuAD answers are either entities or numerics~\cite{squadanalysis} while \ours\ questions can be more open-ended.
% \pl{how many of ours are entities?}.
%points out, the majority of SQuAD answers are either entities or numerics, which is not the case with our data. 
While the average question length (6.5 tokens) is shorter than that of SQuAD (11 tokens), this does not indicate reduced question complexity, as the student (1) cannot access the section to paraphrase it and (2) can be more concise by coreferencing previous interactions.

Figure~\ref{fig:questions} visualizes the most frequent question types in \ours\ based on ``Wh'' words.\footnote{To more effectively visualize sub-boxes like ``what did'', we exclude questions from the tail of the distribution.}  
For a more fine-grained analysis, we randomly sampled 100 questions (each from a different dialog) and manually labeled different phenomena in Table~\ref{tab:question-analysis}.
%:
%\textbf{non-factoid} if the question is not asking about a specific fact;
%\textbf{contextual} if understanding the question requires reading the dialog history;
%\textbf{coreference} if the question contains coreferences to either entities in the article or entities mentioned in the dialog history; and
%\textbf{``anything else''} if the question is asking for addition information about a fact or entity.
%
%
%These properties are not mutually exclusive and Table~\ref{tab:question-analysis} shows some example questions.
Unlike most current QA datasets that focus on factoid questions, our task setup encourages more open-ended questions: about half of \ours\ questions are non-factoid. 
Furthermore, 86\% of questions are contextual, requiring reading the context to resolve coreference; of these, 44\% refer to entities or events in the dialog history, while 61\% refer to the subject of the article.
%,
%containing temporal dependence such as ``next'', ``after''.

\begin{figure*}[t]
    \centering
    \includegraphics[width=\textwidth]{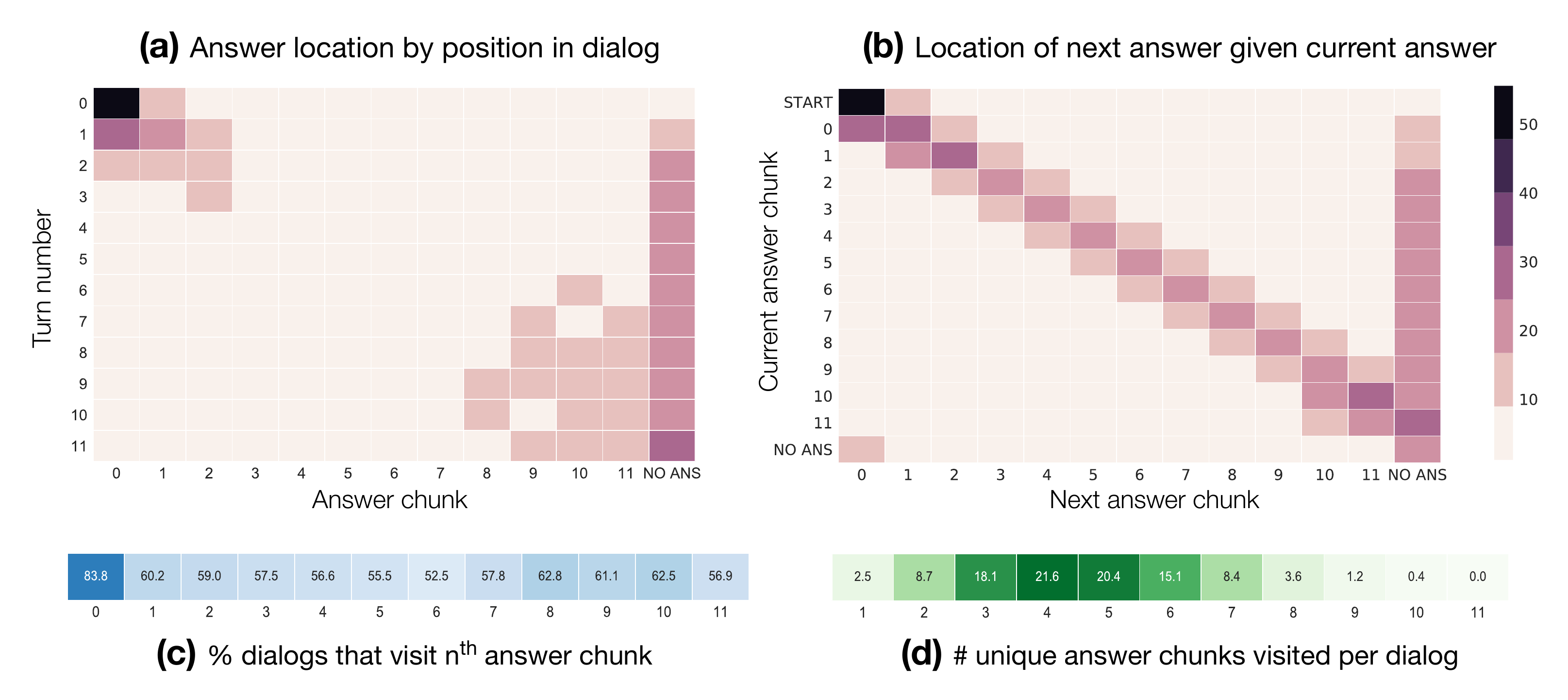}
    \vspace{-5pt}
    \caption{Heatmaps depicting the importance of context in \ours\ dialogs, where (a) and (b) share the same color scale. The student's earlier questions are answered mostly by the first few chunks, while the end of the section is covered in later turns (a). The middle is the least covered portion (c), and dialogs cover around five unique chunks of the section on average (d). The transition matrix (b) shows that the answer to the next question is more likely to be located within a chunk adjacent to the current answer than in one farther away.}
    \label{fig:heatmaps}
        \vspace{-10pt}

\end{figure*}

\paragraph{The role of context}
Dialog context is crucial to understanding and answering \ours\ questions. 
Figure~\ref{fig:heatmaps}a shows that the location of the answer within the text is influenced by the number of questions asked previously.
Early questions are mostly answered in the beginning of the section, while later questions tend to focus on the end of the section.
Interestingly, text in the middle of the section is not asked about as frequently (Figure~\ref{fig:heatmaps}c).
%as text at the beginning or end 
 %and the more 
As more questions get asked, the more likely a question is to be unanswerable.
%We attribute this latter observation to the student narrowing their focus from general to specific questions as the dialog progresses.

Figure~\ref{fig:heatmaps}b shows how the answers progress through different chunks of the evidence text (where each section is divided into 12 chunks of equal size).  The answer to the next question is most frequently either in the same chunk as the previous question or an adjacent chunk, and most dialogs in the dataset cover three to six of the chunks (Figure~\ref{fig:heatmaps}d). 
%\footnote{for this analysis, each section is divided into twelve equally-sized chunks} 
These observations suggest that models for \ours\ must take into account the dialog context. However, results in Section~\ref{sec:experiments} show that solely relying on the location of previous answers is not sufficient.% to answer new questions. %\pl{huh? what other kind of information is there?}

%While the trends in Figure~\ref{fig:heatmaps}ab are strong, \ours\ cannot be solved from just blindly following them, as shown by the low performance of our chunk transition matrix baseline in the next section.

\begin{figure}[t]
    %\vspace{-10pt}
    \centering
    \hspace{-10pt}\includegraphics[width=.5\textwidth]{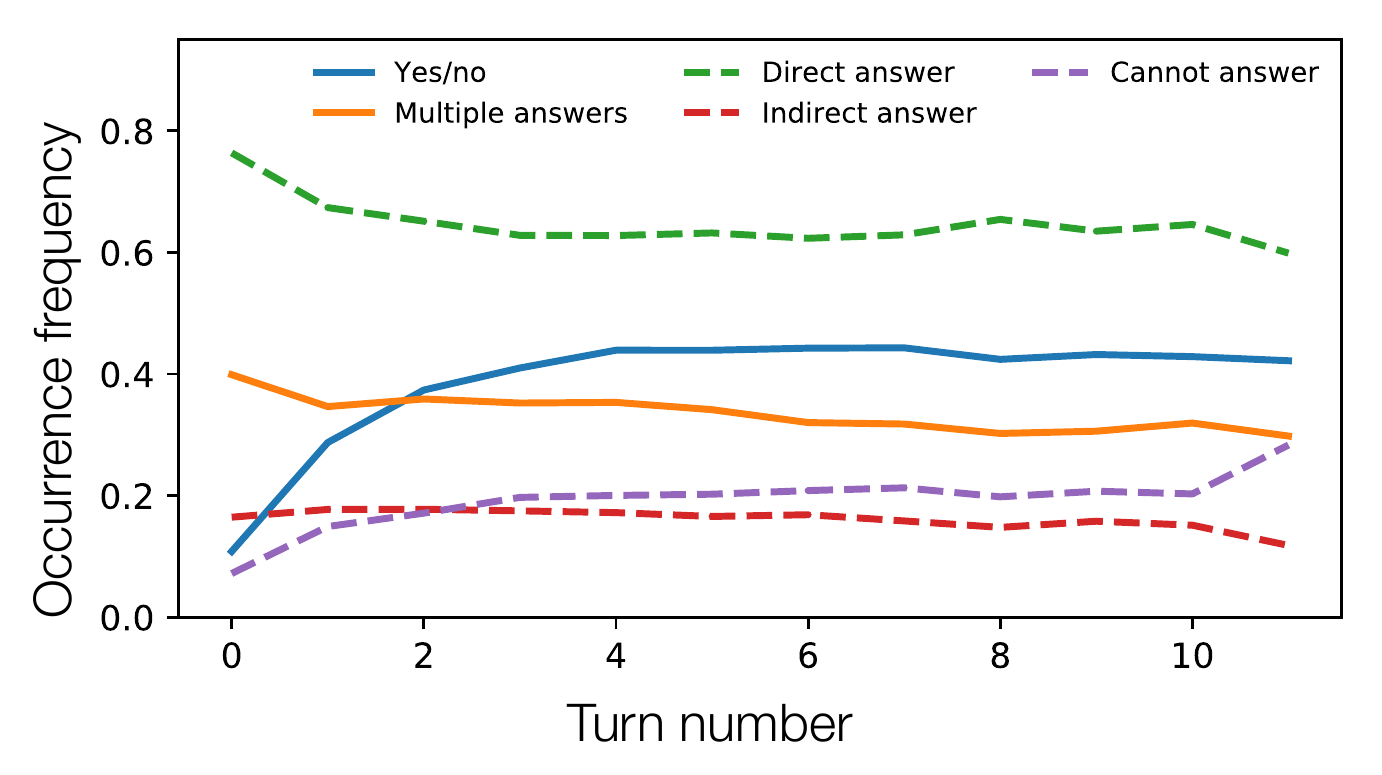}\vspace{-10pt}
    \caption{The number of turns in the dialog influences the student's behavior: they start by asking general questions (i.e., easier to answer, with multiple possible answers) and progress to more specific ones.}
    \label{fig:props}
\end{figure}

Finally, we examine properties of the questions as a function of the turn position in the dialog (Figure~\ref{fig:props}). 
The frequency of yes/no questions increases significantly as the dialogs progress; again, at the beginning of the dialog, students have very little information, so it is harder to formulate a yes/no question. 
The percentage of questions that have multiple answers declines as the dialog progresses, implying students ask general questions first and specific ones later.

\paragraph{Qualitative examples}

\begin{figure}[t]
\small
% \begin{center}
%\colorbox{Goldenrod}
%\begin{mdframed}[style=dialogue]

\begin{tcolorbox}[boxsep=0pt,
                  left=5pt,
                  right=0pt,
                  top=2pt,
                  colback = yellow!5
                  ]%%

    % \hspace{1.25in}
    % \includegraphics[width=.08\textwidth]{figures/daffy_duck.jpg} \\
    \hspace{20pt}{\textbf{Section}: %{\includegraphics[width=.08\textwidth]{figures/pdf/daffyhand.pdf} 
    \href{https://en.wikipedia.org/wiki/Gaelic_Ireland#Invasion}{Gaelic Ireland : Invasion}}\\

\begin{dialogue}

% \end{center}
\vspace{-5mm}
\small

\speak{Student}{\bf What year did the invasion happen?}
\speak{Teacher}\colorbox{CarnationPink!25}{$\hookrightarrow$} %Daffy
{in 1169 the main body of Norman, Welsh and Flemish forces landed in Ireland and quickly retook Leinster and the cities of Waterford and Dublin on behalf of Diarmait.}%, released on April 17, 1937.
%Follow up.

 \speak{Student}{\bf Who was Diarmait?}
\speak{Teacher} \colorbox{CarnationPink!25}{$\bar{\hookrightarrow}$}{King Diarmait Mac Murchada of Leinster.}

\speak{Student}{\bf Where is Leinster located?}
\speak{Teacher}\colorbox{CarnationPink!25}{${\not\hookrightarrow}$} {landed in Ireland and quickly retook Leinster.}

%Follow up? If you want

\speak{Student}{\bf Were invasions common?}
\speak{Teacher}\colorbox{CarnationPink!25}{${\not\hookrightarrow}$}\colorbox{blue!15}{No answer}

%Follow up.

\speak{Student}{\bf Are there any other interesting aspects about this article? }
\speak{Teacher}\colorbox{CarnationPink!25}{ $\bar{\hookrightarrow}$}
\colorbox{red!25}{Yes,} IPope Adrian IV, the only English pope, had already issued a Papal Bull in 1155 giving Henry II of England authority to invade Ireland.

%Follow up, multiple answers

\speak{Student}{\bf Who lead the invasion? }
\speak{Teacher}\colorbox{CarnationPink!25}{${\not\hookrightarrow}$}\colorbox{blue!15}{No answer}

\speak{Student}{\bf Did England defeat the Irish armies?  }
\speak{Teacher}\colorbox{CarnationPink!25}{${\not\hookrightarrow}$}\colorbox{blue!15}{No answer}

\end{dialogue}
\end{tcolorbox}
\caption{A less successful dialog from \ours. The student struggles to get information despite asking good questions. The teacher attempts to provide extra context to guide the student, but the dialog ultimately ends because of too many unanswerable questions.} 
\label{fig:exdial3}
\end{figure}

Figures~\ref{fig:exdial2} and ~\ref{fig:exdial3} contain two representative dialogs from \ours. 
Longer dialogs sometimes switch topics (such as in Figure~\ref{fig:exdial2} about ``academic work'') and often go from general to specific questions.   
Students whose questions go unanswered commonly resort to asking their teacher for any interesting content; even if this strategy fails to prolong the dialog as in Figure~\ref{fig:exdial3}, models can still use the dialog to learn when to give no answer.

%\paragraph{Validation Answers}
% \begin{figure}[t]
%     \centering
%     \includegraphics[width=.5\textwidth]{figures/pdf/transition_heatmap.pdf}
%     \caption{Transition matrix}
%     \label{fig:transition_heatmap}
% \end{figure}

% \begin{figure}[t]
%     \centering
%     \includegraphics[width=.5\textwidth]{figures/pdf/pos_heatmap.pdf}
%     \caption{Answer by position heatmap}
%     \label{fig:pos_heatmap}
% \end{figure}

\begin{table}[t]
 \centering
 {\footnotesize

 \setlength\tabcolsep{0.0ex}
 \begin{tabular}{L{1.5cm}C{0.8cm}L{5.5cm}}
 \toprule
 Question type & \% & Example\\
 \midrule
     Non-factoid & 54 &
     Q: \textbf{Were} the peace talks \textbf{a success?} 
     \newline Q: \textbf{What was} her childhood \textbf{like?}
     %\newline A: After several days of unfruitful negotiations, without reaching any specific agreements, Maldonado proposed an indefinite suspension of hostilities.
     \\
\midrule
     Contextual & 86 & 
\\
    % \cmidrule(){3-3} 
    \cmidrule(){3-3}
     \hspace{10pt}Coref \newline \hspace{6pt}(article) & 61 & 
     Title: Paul C\'ezanne: Early years\newline
     Q: When did \textbf{he} start painting?
     \\
         \cmidrule(){3-3}

     \hspace{10pt}Coref \newline \hspace{6pt}(history) & 44 &
     Q: What was special about the Harrah's? \newline
     A: project was built by Trump with financing from the Holiday Corporation. \newline
     Q: \textbf{Which led to what?}
     %\newline
     \\
     
     \cmidrule(){3-3} 
     %\midrule
   
     \hspace{6pt}Anything \newline \hspace{14pt}else? & 11 & 
     Q: What \textbf{other acting} did he do?
     \newline Q: \textbf{What else} did he research?
\\
 \bottomrule
 \end{tabular}
 }
 \caption{\label{tab:question-analysis}
 An analysis of \ours\ questions. {\bf Non-factoid} questions do not ask about specific facts, while
 {\bf contextual} questions require reading the history to resolve coreferences to the dialog history and/or article.
 %previously mentioned information.
 %; {\bf coreference}: either to entities in the article or dialog history; and {\bf open ended}: a query for addition information.
 }
\end{table}

\section{Experimental Setup}
\label{sec:setup}

% In this section, we describe our task setup and introduce evaluation metrics on \ours. \lzcomment{cut this sentence?}

%\subsection{Task}
We consider the following QA task: given the first $k$ questions and $k$ ground-truth answers in the dialog, all supporting material (entity $e$, topic $t$, background $b$, and section text $s$), and question $q_{k+1}$, we predict the answer span indices $i,j$ in the section text $s$. 
Since affirmation questions are incomplete without a yes/no answer and the continuation feedback is important for information-seeking dialog, we predict the dialog acts $v$, which with the span form the final answer prediction $a_{k+1}$. 
% that answer the question \pl{reads like dialog act indicators answer the question, when I assume you mean spans too; also I still don't really understand what dialog act indicators are or why they're useful except for yes/no/don't know; should explain that more}.

All of our experiments are carried out on a train/dev/test split of 83.5k/7.3k/7.3k questions/answer pairs, where no sections
% \pl{what are sections? articles?}
are shared between the different folds. Questions in the training set have one reference answer, while dev and test questions have five references each.
% \pl{before you used 'annotation'; be consistent; I'd just say 'annotated answer' or 'reference answer' to be unambiguous} 
For all experiments, we do not evaluate on questions with a human F1 lower than 40, which eliminates roughly 10\% of our noisiest annotations.\footnote{A manual inspection of annotations below this threshold revealed many lower quality questions; however, we also report unthresholded F1 in the final column of Table~\ref{tab:experiments}.}
% \eccomment{but we have column on this in table 4?}
% \pl{shouldn't these things be part of the dataset description earlier?}

\subsection{Evaluation Metrics}
\label{subsec:eval}
%For evaluation purposes, we collect between four alternative answers to questions. 
Our core evaluation metric, word-level F1, is implemented similarly to SQuAD~\cite{rajpurkar2016squad}: precision and recall are computed by considering the portion of words in the prediction and references that overlap after removing stopwords.\footnote{Since our answer spans have vaguer boundaries than the shorter ones in SQuAD, exact match is not a useful metric.} For \texttt{no answer} questions, we give the system an F1 of one if it correctly predicts \texttt{no answer} and zero otherwise.\footnote{Because the validation task was more susceptible to spam by constant annotation of ``no-answer,'' we only allow ``no-answer'' if the majority of references marked ``no-answer'', removing other answers. If ``no-answer'' is not the majority answer, we remove all instances of ``no-answer''.}
Like SQuAD, we compute the maximum F1 among all references; however, since many \ours\ questions have multiple valid answers, this metric varies significantly with the number of reference annotations.
To make oracle human and system performance comparable, given $n$ references, we report the average of the maximum F1 computed from each $n-1$ subset with respect to the heldout reference. 
% of the maximum F1 with respect to any of the $n-1$ and heldout reference. 
% \pl{you mean you take the average over all $n-1$ subsets of the maximum F1 with respect to any of the $n-1$ and heldout reference?}

Additionally, since averaged F1 can be misleading for questions with multiple valid answers, we introduce the human equivalence score (HEQ), a performance measure for judging whether a system's output is as good as that of an average human.\footnote{In cases with lower human agreement on F1, if a system produces one reference exactly (F1 = 100), it will get points that it can use to offset poor performance on other examples.}
HEQ measures the percentage of examples for which system F1 exceeds or matches human F1.
%To compute HEQ, we check if the system's F1 matches or exceeds human F1 on a particular question.
We compute two variants: (1) the percentage of questions for which this is true (HEQ-Q), and (2) the percentage of dialogs for which this is true for every question in the dialog (HEQ-D).
A system that achieves a value of 100 on HEQ-D can by definition maintain average human quality output over full dialogs. 

For dialog acts, we report accuracy with respect to the majority annotation, breaking ties randomly.

\section{Experiments}
\label{sec:experiments}

%In this section, we describe and analyze the results of a variety of baselines, sanity checks, and upper bounds.

\subsection{Sanity checks}
% We compute three naive baselines for \ours~. 
%\begin{itemize}
%    \item 
    \paragraph{\bf Random sentence} This baseline selects a random sentence in the section text $s$ as the answer (including \texttt{no answer}).% It does not output any other dialog acts. 
    
%    \item 
    \paragraph{\bf Majority} The majority answer outputs \texttt{no answer} and the majority class for all other dialog acts (\texttt{neither} for affirmation and \texttt{don't follow up} for continuation).
    
%    \item 
    \paragraph{\bf Transition matrix} We divide the supporting text into 12 chunks (with a special chunk for \texttt{no answer}) and use the transition matrix (computed from the training set) in  Figure~\ref{fig:heatmaps}b to select an answer given the position of the previous answer. This baseline does not output other dialog acts.
%\end{itemize}

\subsection{Upper bounds}
% We compute three upper bounds for \ours.
   
    \paragraph{Gold NA + TM} This is the same transition matrix (TM) baseline as before, except that for questions whose gold annotations are \texttt{no answer}, we always output \texttt{no answer}.
    
    \paragraph{Gold sentence + NA} To see if \ours\ can be treated as an answer sentence selection problem, we output the sentence from $s$ with the maximal F1 with respect to references, or \texttt{no answer} for unanswerable questions.
    
    \paragraph{Human performance} We pick one reference as a system output and compute the F1 with respect to the remaining references using the method described in Section~\ref{subsec:eval}. By definition, all HEQ measures are 100, and we report agreement for the affirmation dialog act.\footnote{We did not collect multiple annotations for the continuation dialog act and so omit it.}
    
\subsection{Baselines}
    \paragraph{Pretrained InferSent} To test the importance of lexical matching in our dataset, we output the sentence in $s$ whose pretrained InferSent representation~\cite{conneau2017supervised} has the highest cosine similarity to that of the question.
    
    \paragraph{Feature-rich logistic regression} We train a logistic regression using Vowpal Wabbit~\cite{langford2007vowpal} to select answer sentences. We use simple matching features (e.g., n-gram overlap between questions and candidate answers), bias features (position and length of a candidate), and contextual features (e.g., matching features computed with previous questions / answers, turn number).
    
    % \paragraph{InferSent Classifier} The same as above except a logistic regression was trained with InferSent representations as features.
    
    \paragraph{BiDAF++} We use a re-implementation of a top-performing SQuAD model~\cite{peters2018deep} that augments bidirectional attention flow~\cite[BiDAF]{seo2016bidirectional} with self-attention~\cite{clark2017simple} and contextualized embeddings.\footnote{The AllenNLP~\cite{Gardner2017AllenNLP} implementation we use reaches 82.7 on the SQuAD development set, compared to the paper's reported 85.8 on SQuAD; regardless, this implementation would have been state-of-the-art less than a year ago, making it an extremely strong baseline.}

    A token for \texttt{no answer} is appended to $s$ to enable its prediction following~\citet{Levy2017ZeroShotRE}. Additionally, we modify the model for our task to also predict dialog acts, placing a classifier over the same representation used to predict the end position of the predicted span. 
    % At train time, we use representations corresponding to the ground-truth span ending position, while at test-time dialogs acts are predicted from the end position corresponding to the model's predicted span.
    
    % All BiDAF++ models were trained using AllenNLP~\cite{Gardner2017AllenNLP}, using standard settings, with dropout .2,  ADAM\cite{} for 20 epochs and learning rate decay when validation loss doesn't improve. 
    
    \paragraph{BiDAF++ w/ k-ctx}
    As BiDAF++ does not model any dialog context, we modify the passage and question embedding processes to consider the dialog history. We consider context from the previous \emph{k} QA pairs.\footnote{Our implementation is available in AllenNLP.}
    \begin{itemize}
	\item \textbf{Passage embedding} We explicitly identify the previous $k$ answers within the section text by  concatenating marker embeddings to the existing word embeddings. 
	\item \textbf{Question embedding}  Naively prepending the previous \emph{k} questions to the current question did not show gains in initial experiments. We opt instead to simply encode the dialog turn number within the question embedding. 
    \end{itemize}

\begin{table*}[t]
    \centering
    \begin{tabular}{l|c|c|c|c|c|c}
    \toprule
     & F1 & HEQ-Q & HEQ-D & Yes / No & Follow up & F1 (All) \\
    \toprule
    Random sentence & 15.7 / 15.6 & 6.9 / 6.9  & 0.0 / 0.1 & --- &--- &16.4 / 16.3 \\
    Majority answer & 22.7 / 22.5 & 22.7 / 22.5 & 0.5 / 0.4 & 78.8 / 77.6 &57.9 / 56.7 &20.2 / 20.0\\
    Trans. matrix (TM) & 31.8 / 31.5 & 15.8 / 15.8 & 0.1 / 0.2 & --- &  --- &31.2 / 30.9\\
    \midrule
    Pretrained InferSent &  21.4 / 20.8 & 10.2 / 10.0 & 0.0 / 0.0 &--- & --- & 22.0 / 21.4 \\
    Logistic regression & 34.3 / 33.9 & 22.4 / 22.2 & 0.6 / 0.2 & --- &--- & 34.3 / 33.8 \\
    BiDAF++ (no ctx) & 51.8 / 50.2 & 45.3 / 43.3 & 2.0 / 2.2 & 86.4 / 85.4 &59.7 / 59.0 & 50.1 / 48.2%51.6 / 49.8 & 44.8 / 42.5 & 2.3 & 1.6 &  % 52.0 / 50.1& 45.2 / 42.9 & 2.0 / 1.2 & {86.9} / 86.4 & 60.5 / 59.9& 50.1 / 48.1  \\%47.0 / 45.7  & 38.8 / 37.1 & 0.9 / 1.9 & 85.4 / 84.9 & 44.6 / 43.4 &  45.0 / 43.7  \\52.0 / 50.4& 45.5 / 43.5 & 2.4 / 2.5 & 86.5 / 86.8 & 49.8 / 48.2
    \\
    BiDAF++ (w/ 1-ctx) &59.9 / 59.0 & 54.9 / 53.6 &4.7 / 3.4& 86.5 / \textbf{86.1} &61.3 / 60.3 &57.5 / 56.5 % 59.1 / 58.9 & 54.0 / 53.3 & 4.0 / 3.3 & {87.3} / {86.9} & 61.7 / 60.7 & 56.9 / 56.5\\% 55.4 / 55.4 & 49.1 / 48.9 & 2.5 / 2.3 & 85.0 / 85.4 & 45.3 / 43.8 & 53.2 / 53.1 \\
    \\
    BiDAF++ (w/ 2-ctx) &\textbf{60.6} /\textbf{ 60.1} &\textbf{55.7} / \textbf{54.8} & \textbf{5.3} / 4.0 & \textbf{86.6} / 85.7 & \textbf{61.6} /\textbf{ 61.3} & \textbf{58.3 / 57.8} %\textbf{61.5} / \textbf{60.2} & \textbf{57.0} / \textbf{55.1} & \textbf{5.2} / \textbf{4.3} & \textbf{87.7} / \textbf{87.3} & {62.0} / \textbf{61.1} & \textbf{59.1} / \textbf{57.7}\\ %60.2 / \textbf{59.8} & 55.1 / \textbf{55.0}& \textbf{4.8} / \textbf{4.3} & \textbf{86.9} / 86.8 & 58.2 / \textbf{57.6} \\% 57.9 / {\bf 57.2} & 52.3 / {\bf 51.3} & {\bf 2.8} / 3.7 & 85.9 / 85.1 & 44.4 / 43.4 & 55.3 / 54.5 \\
    \\
    BiDAF++ (w/ 3-ctx) &\textbf{60.6} / 59.5 & 55.6 / 54.5 & 5.0 / \textbf{4.1} &86.1 / 85.7 &\textbf{61.6} / 61.2 &58.1 / 57.0%{60.4} / 59.4 & {55.8} / 54.4 & 4.1 / 3.6 & {87.2} / 86.4 & \textbf{62.4 }/ 61.0 &{58.4} / 57.0\\%{\bf 58.2} / 57.1 & {\bf 52.4} / 51.1  & 2.7 / {\bf 4.0} & {\bf 86.4} / {\bf 86.2}  & {\bf 45.5} / {\bf 43.7} & {\bf 55.8} / {\bf 54.7}  \\
    \\
    \midrule
    Gold NA + TM & 43.0 / 42.6 & 27.4 / 27.4 & 1.0 / 0.8 & --- &  --- &41.0 / 40.6\\
    %BiDAF++ (gold paragraph) &  &  & & \\
    Gold sentence + NA & 72.4 / 72.7 & 61.8 / 62.7 & 9.8 / 9.7& --- & --- & 70.8 / 71.2\\
    Human performance & 80.8 / 81.1  & 100 / 100  & 100 / 100 & 89.4 / 89.0 &  --- &74.6 / 74.7 \\  \bottomrule
    \end{tabular}
             \vspace{-5pt}

    \caption{Experimental results of sanity checks (top), baselines (middle) and upper bounds (bottom) on \ours. Simple text matching baselines perform poorly, while models that incorporate the dialog context significantly outperform those that do not. Humans outperform our best model by a large margin, indicating  room for future improvement.}
    \label{tab:experiments}
         \vspace{-5pt}

\end{table*}

\subsection{Results}
Table~\ref{tab:experiments} summarizes our results (each cell displays dev/test scores), where dialog acts are Yes/No (affirmation) and Follow up (continuation). 
For comparison to other datasets, we report F1 without filtering low-agreement QA pairs (F1').  
        
\paragraph{Sanity check} Overall, the poor sanity check results imply that \ours\ is very challenging. Of these, following the transition matrix (TM) gives the best performance, reinforcing the observation that the dialog context plays a significant role in the task.
%~\ours.

\paragraph{Upper bounds}The human upper bound (80.8 F1) demonstrates high agreement. %between references.
While Gold sentence + NA does perform well, indicating that significant progress can be made by treating the problem as answer sentence selection, HEQ measures show that span-based approaches will be needed achieve average human equivalence.
Finally, the Gold NA + TM shows that \ours\ cannot be solved by ignoring question and answer text.

\paragraph{Baselines} Text similarity methods such as bag-of-ngrams overlap and InferSent are largely ineffective on ~\ours, which shows that questions have little direct overlap with their answers.
On the other hand, BiDAF++ models make significant progress, demonstrating that existing models can already capture a significant portion of phenomena in ~\ours. 
The addition of information from previous turns (w/ 1-ctx) helps significantly, indicating that integration of context is essential to solving the task.
While increasing the context size in BiDAF++ continues to help, we observe saturation using contexts of length 3, suggesting that more sophisticated models are necessary to take full advantage of the context. 
% \pl{is that the conclusion, or that you don't need context more than 3?}
Finally, even our best model underperforms humans: the system achieves human equivalence on only 60\% of questions and 5\% of full dialogs.

\begin{figure*}[t]
    \centering\vspace{-5pt}
    \includegraphics[width=1.0\textwidth]{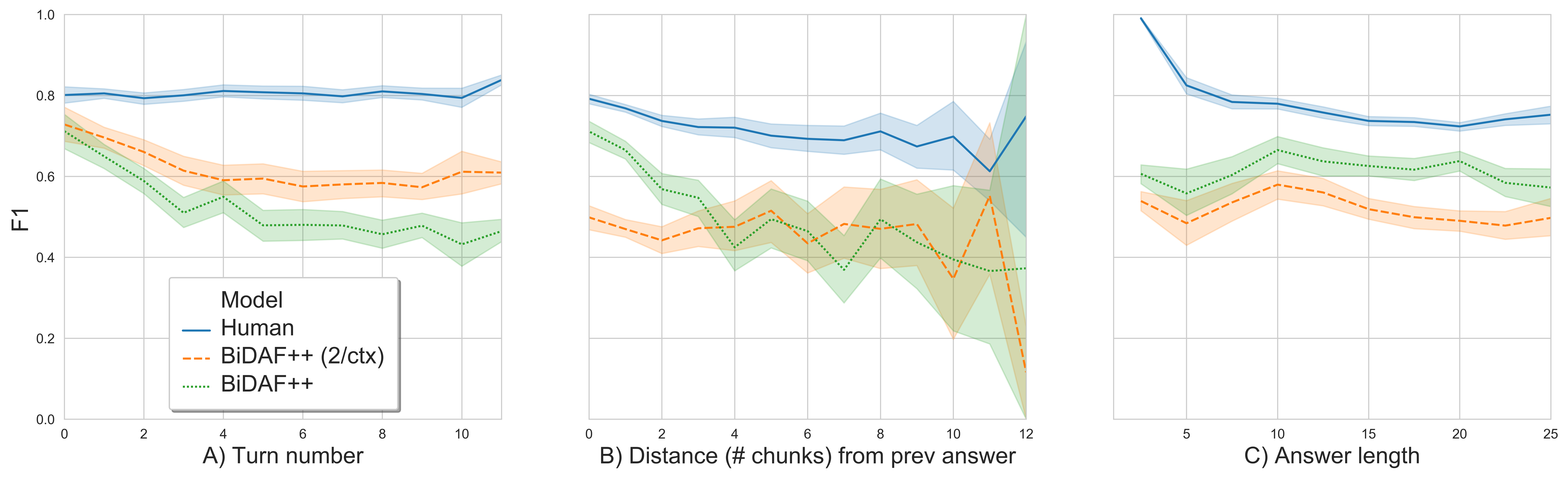}\vspace{-5pt}
    \caption{The F1 scores of baseline models and human agreements based on dialog turn number, answer's distance from previous answer, and the answer span token length.}
    \label{fig:f1_trend}
     \vspace{-10pt}
\end{figure*}

\subsection{Error Analysis}
In this section, we analyze the development set performance of our best context-aware model (BiDAF++ w/ 2-ctx), our best context-agnostic model (BiDAF++), and humans.  Figure~\ref{fig:f1_trend} contains three plots showing how F1 scores of baseline models and human agreement vary with (1) turn number, (2) distance from previous answer,\footnote{We divide the text into 12 equally-sized chunks and compute the difference of the current and previous chunk indices.}
% \footnote{Measured by the difference between the chunk index in section text (12 equally sized chunks) of the current and previous answer span start token, excluding pairs involving unanswerable questions.}
and (3) answer length in tokens. Taken as a whole, our analysis reveals significant qualitative differences between our context-aware and context-agnostic models beyond simply F1; additionally, human behavior differs from that of both models.
%The first plot shows how performance varies as dialog proceeds. 

In the first plot, human agreement is unchanged throughout the dialog while the performance of both models decreases as the number of turns increases, although the context-aware model degrades less. 
While continuing a dialog for more turns does not affect human agreement, the second plot shows that human disagreement increases as the distance between the current answer's location within the section text and that of the previous answer increases. Larger distances indicate shifts in the student's line of questioning (e.g., if the teacher told the student not to follow up on the previous question). The plot also shows that model performance suffers (significantly more than humans) as distance increases, although the context-aware model can tolerate smaller shifts better than the context-agnostic model. In the last plot, human agreement is higher when the answer span is short; in contrast, our model struggles to pin down short answers compared to longer ones. 

%When the answer is in proximity to the previous answer, which is marked in the passage, the context sensitive model significantly outperform the context insensitive model.
%When the distance is longer, the context sensitive model actually does not outperform context insensitive model, questioning whether it exploits context more than searching its proximity to prior answer. 
%The last plot shows whether the answer length %(number of tokens in the validation answer %string, excluding unanswerable questions) %impacts the performances. 

The plots demonstrate the increased robustness of the context-aware model compared to BiDAF++.  This finding is reinforced by examining the difference in model performance on questions where previously the teacher recommended the student to ``follow up'' vs. not to follow up. 
The context-aware baseline performs 6 HEQ-Q higher on the ``follow up" questions; in contrast, the context-agnostic baseline shows no HEQ-Q difference between the two types of questions. This discrepancy stems from the context-agnostic model's inability to take advantage of the location of the previous answer.

%performance difference based on the continuation of the previous query. 

%Correctly identifying unanswerable questions is easier for baselines than identifying the correct answer spans: HEQ-Q on unanswerable questions was 61.6, while 55.6 on the others.
%Also, Yes/No questions (48.1/54.0 HEQ) are harder than questions without Yes/No property (57.8 HEQ). 
% 0.518679
% 2    0.574646

% \begin{figure}[t]
%     \centering
%     \includegraphics[width=0.5\textwidth]{figures/filename.png}
%     \caption{Human equivalence score based on dialogue acts of the previous question.}
%     \label{fig:heq_trend}
% \end{figure}

%answerable 0.556123
%not-answerable 0.616420

%0    n
%1    x
%2    y
%Name: Yes/No, dtype: object 0    0.481132
%1    0.578826
%2    0.540094 

% 0    m
% 1    n
% 2    y
% Name: Prev Followup, dtype: object 0    0.540196
% 1    0.518679
% 2    0.574646
% Name: HEQ, dtype: float64
\section{Related Work}\label{sec:related}
 \vspace{-2pt}
\paragraph{Reading Comprehension}
Our work builds on span based reading comprehension~\cite{rajpurkar2016squad, JoshiTriviaQA2017, trischler2016newsqa}, while also incorporating innovations such as curating questions independently of supporting text to reduce trivial lexical overlap~\cite{JoshiTriviaQA2017,Kocisk2017TheNR} and allowing for unanswerable questions~\cite{trischler2016newsqa,Rajpurkar2018KnowWY}.
%Our questions are collected independently from the evidence document, %as the student can only see the section title and a short summary. 
%This can prevent potential question bias, as observed in other recent work
%eliminating a commonly recognized source of bias~\cite{JoshiTriviaQA2017,trischler2016newsqa,Kocisk2017TheNR}. 
%Our dataset contains a significant portion of naturally occurring unanswerable question (20.2\%), which is missing in most other datasets. %which provides a more realistic scenario for question answering. 
%Unlike most dataset which exclusively deals with factoid questions, 
We handle open-ended questions like in MSMARCO~\cite{Nguyen2016MSMA}, with multiple references, but we are the first to incorporate these into information-seeking dialog.
 \vspace{-2pt}
\paragraph{Sequential QA}
Our work is similar to sequential question answering against knowledge bases~\cite{IyyerSQA2017} and the web~\cite{talmor2018web}, but instead of decomposing a single question into smaller questions, we rely on the curiosity of the student to generate a sequence of questions. % and aim to decompose complex question into simpler questions.
%In contrast, questions in \ours\ do not decompose a specific information need.
Such open information seeking was studied in semantic parsing on knowledge bases~\cite{Dahl1994ExpandingTS} and more recently with modern approaches~\cite{saha2018complex}, but with questions paraphrased from templates. 
Concurrent to our work,~\citet{Saeidi2018QuARC} proposed a task of generating and answering yes/no questions for rule focused text (such as traffic laws) by interacting with a user through dialog.
Also concurrently, ~\newcite{Reddy18} propose conversational question answering (CoQA) from text but allow both students and questioners to see the evidence. As a result, a large percentage of CoQA answers are named entities or short noun phrases, much like those in SQuAD. In contrast, the asymmetric nature of \ours\ forces students to ask more exploratory questions whose answers can be potentially be followed up on.\footnote{On average, CoQA answers are 2.7 tokens long, while SQuAD's are 3.2 tokens and \ours\ 's are over 14 tokens.}

%modeling regulatory yes/no questions that need follow-up information from user.
%Also concurrently, ~\newcite{Reddy18} propose conversational question answering (CoQA) from text but allow both students and teachers to see the evidence. As a result, a large percentage of CoQA answers are named entities or short noun phrases, much like those in SQuAD. In contrast, the asymmetric nature of \ours\ forces students to ask more exploratory questions whose answers can be potentially be followed up on.\footnote{On average, CoQA answers are 2.7 tokens long, while SQuAD's are 3.2 tokens and \ours\ 's are over 14 tokens.} 

%In \ours , new questions arise exclusively based on the prior interactions.
%This discourage students from asking for specific named entities or short noun phrases which might be absent (from what?), and encourages teacher to convey more information in its answers.

%, which results in more SQuAD-like interactions.

%\eunsol{add a sentence or two}

%Information seeking, conversational dialog on databases was considered CSQA~\cite{saha2018complex}, but their questions are paraphrased from templates. 
 \vspace{-2pt}
\paragraph{Dialog}
\ours~fits into an increasing interest in open domain dialog, mostly studied in the context of social chit-chat~\cite{li2016deep,Ritter2011DataDrivenRG,fang2017sounding,ghazvininejad2017knowledge}.
Most related to our effort is visual dialog~\cite{das2017visual}, which relies on images as evidence instead of text. 
More explicit goal driven scenarios, such as bargaining~\cite{lewis2017deal} and item guessing~\cite{he2017learning} have also been explored, but the language is more constrained than in ~\ours. Information-seeking dialog specifically was studied in~\newcite{stede2004information}.

%\pl{writing is a bit choppy here}

%While allowing free form answers makes it easier to continue dialogs, restricting the answer to be span-based significantly increases the ease of the evaluation. 

%With the exception of NewsQA dataset, which crowdsourced questions from news article summaries, no other dataset contains naturally occuring unanswerable questions. 

% \subsection{Comparison to other datasets}
% \textbf{Advantages over other datasets}
% \begin{itemize}
%     \item Multi-turn (\ours, SQA, CSQA~\cite{saha2018complex})
%     \item Scale (\ours, SQUAD, triviaQA, narrativeQA, MSMARCO, NewsQA)
%     \item Cannot answer (\ours, MSMARCO, NewsQA)
%     \item Multiple answer (\ours, MSMARCO)
%     \item Open ended (\ours, NewsQA (?), MSMARCO)
%     \item Low importance of string matching (\ours)
%     \item Natural topics/questions
%     \item Ease of evaluation (\ours, SQUAD, triviaQA, NewsQA)
% \end{itemize}

% 

\section{Conclusion}
\label{sec:conclusion}

In this paper, we introduce \ours, a large scale dataset of information-seeking dialogs over sections from Wikipedia articles. Our data collection process, which takes the form of a teacher-student interaction between two crowd workers, encourages questions that are highly contextual, open-ended, and even unanswerable from the text. Our baselines, which include top performers on existing machine comprehension datasets, significantly underperform humans on \ours.
We hope this discrepancy will spur the development of machines that can more effectively participate in information seeking dialog.
 %we hypothesize that improvements on the dataset will come from advances in modeling the dialog context.
\section*{Acknowledgments}
\label{sec:acknowledgments}

\ours\ was jointly funded by the Allen Institute for Artificial Intelligence and the DARPA CwC program through ARO (W911NF-15-1-0543). 
We would like to thank anonymous reviewers and Hsin-Yuan Huang who helped improve the draft. 

\bibliographystyle{bib/acl_natbib_nourl}
\bibliography{bib/journal-full,bib/main}

\end{document}